\documentclass[10pt,twocolumn,letterpaper]{article}

\usepackage{iccv}
\usepackage{times}
\usepackage{epsfig}
\usepackage{graphicx}
\usepackage{amsmath}
\usepackage{amssymb}

\usepackage[breaklinks=true,bookmarks=false]{hyperref}

\iccvfinalcopy 

\def\iccvPaperID{****} 

\begin{document}

\title{End-to-End Denoising of Dark Burst Images Using Recurrent Fully Convolutional Networks}

%

\author{Di Zhao\thanks{They contributed equally to this work.}  ~\thanks{Fuzhou University. This work was done during an internship at TCL.}\\
{\tt\small pldbe1@163.com}
\and
Lan Ma\footnotemark[1] ~\thanks{TCL Corporate Research}\\
{\tt\small malantju@gmail.com}\\
\and
Songnan Li\footnotemark[3]\\
{\tt\small lisongnan1982@gmail.com}
\and
Dahai Yu\footnotemark[3]\\
{\tt\small dahai.yu@tcl.com}
}

\maketitle

\begin{abstract}
When taking photos in dim-light environments, due to the small amount of light entering, the shot images are usually extremely dark, with a great deal of noise, and the color cannot reflect real-world color. Under this condition, the traditional methods used for single image denoising have always failed to be effective. One common idea is to take multiple frames of the same scene to enhance the signal-to-noise ratio. This paper proposes a recurrent fully convolutional network (RFCN) to process burst photos taken under extremely low-light conditions, and to obtain denoised images with improved brightness. Our model maps raw burst images directly to sRGB outputs, either to produce a best image or to generate a multi-frame denoised image sequence. This process has proven to be capable of accomplishing the low-level task of denoising, as well as the high-level task of color correction and enhancement, all of which is end-to-end processing through our network. Our method has achieved better results than state-of-the-art methods. In addition, we have applied the model trained by one type of camera without fine-tuning on photos captured by different cameras and have obtained similar end-to-end enhancements.

\end{abstract}


\section{Introduction}
In recent years, consumer electronics products have been changing with each passing day and, as a result of this rapid and steady increase in available technology and customers’ expectations, smartphones are essentially necessary for everyone~\cite{decker2010estimating} . People are becoming accustomed to using mobile phones instead of professional cameras to take pictures in various environments because of the portability of mobile phones. One of the special photo scenarios that was once reserved for professional cameras, but is increasingly important to non-professional photographers, occurs when smartphone users take pictures in extremely dark environments, usually under 3 lux~\cite{hasinoff2016burst}. For example, when taking a photo under the light of the full moon, the light is usually about 0.6 Lux. Another example, which is even more common among amateur photographers, comes when users take a photo in a dark indoor environment without lights (in which the light level is roughly 0.1 lux). We are interested in this kind of extremely dark scene because in this kind of dim environment, taking pictures with a portable mobile phone can help us ``see'' things that are difficult to see with the naked eye.

Compared with single-lens reflex (SLR) cameras, using a mobile phone to get good pictures in this kind of dark environment is extremely difficult. This is due to the fact that smartphones, by their very design, preclude the possibility of large aperture lens designs, which makes it impossible to collect enough light when taking pictures. Due to the small amount of light entering the aperture, the shot images are usually extremely dark with a great deal of noise; furthermore, the color in the photos cannot reflect the real-world color of the image~\cite{remez2017deep}. 

Theoretical deduction has proven that increasing the photon counts can effectively improve the image signal-to-noise ratio (SNR)~\cite{el2005cmos}. There are many ways to increase the photon counts, one of which is to increase the exposure time~\cite{xiao2009mobile}. However, this is usually done by mounting the camera on a tripod. During the long exposure time, any movement of the camera and any moving objects in the scene will cause objects to become blurry in the picture.  An additional problem of increasing the exposure time occurs when one takes photos of high dynamic range scenes; the darkest areas of the image will likely still have a lot of noise, while the brightest areas will tend to be saturated~\cite{seetzen2004high}. Another way to increase the exposure time in the industry is through the multi-frame fusion method which combines many short exposure frames together. It is equivalent to increasing the exposure time to achieve the purpose of improving the SNR~\cite{hasinoff2016burst}. 

Traditional denoising methods based on single-frame images have matured and the performance is slowly approaching saturation~\cite{ghimpecteanu2016decomposition}\cite{irum2015review}\cite{jain2016survey}. Furthermore, in extremely dark scenes, all the current traditional denoising methods have failed to be effective~\cite{plotz2017benchmarking}\cite{chen2018learning}. Chen, et. al., has proven that a fully-convolutional network, which operates directly on single-frame raw data can replace the traditional image processing pipeline and more effectively improve the image quality~\cite{chen2018learning}.  However, after further investigation, it turns out that in a great deal of dark cases, using Chen, et. al.'s network processing on single-frame results may miss a lot of details; additionally, sometimes the color cannot reflect the real color. This is because, under dark environments, the SNR of the shot's image is quite low. A great deal of useful information will be concealed by strong noise and cannot be fully recovered through a single image.

Inspired by this work and traditional multi-frame denoising methods, we propose a Recurrent Fully Convolutional Network (RFCN) to process burst photos taken under extremely low-light conditions and to obtain denoised images with improved brightness. The major contributions of the work can be summarized as follows:

1.	We proposed an innovative framework which directly maps multi-frame raw images to denoised and color-enhanced sRGB images, all of which is end-to-end processing through our network. By using raw data, we were able to avoid the information loss which occurs in the traditional image processing pipeline. We have established that using raw bursts images achieves better results than state-of-the-art methods under dark environments.

2.	We have proven, moreover, that our framework has high portability and cross-platform potential, i.e., the model trained by one mobile phone can be directly applied to different cameras' raw bursts without fine-tuning and can obtain a similar level of enhancement. 

3.	Our framework is relatively flexible since we can produce either a best image or generate a multi-frame denoised image sequence. This opens up the possibility of expanding our framework to cover video denoising, as well. 

The paper is organized as follows: The first part consists of an introduction of the problems that occur when photos are taken under dark environments. The second part gives a general overview of image denoising methods and some related work. The third part,``Methods and Materials,'' describes the overall framework, the network architecture and training details, and the data used to train and test the network. Finally, the results, discussion, and conclusion are detailed in Sections 4 and 5.

\begin{figure*}[t]
\centering
\includegraphics[width=16cm]{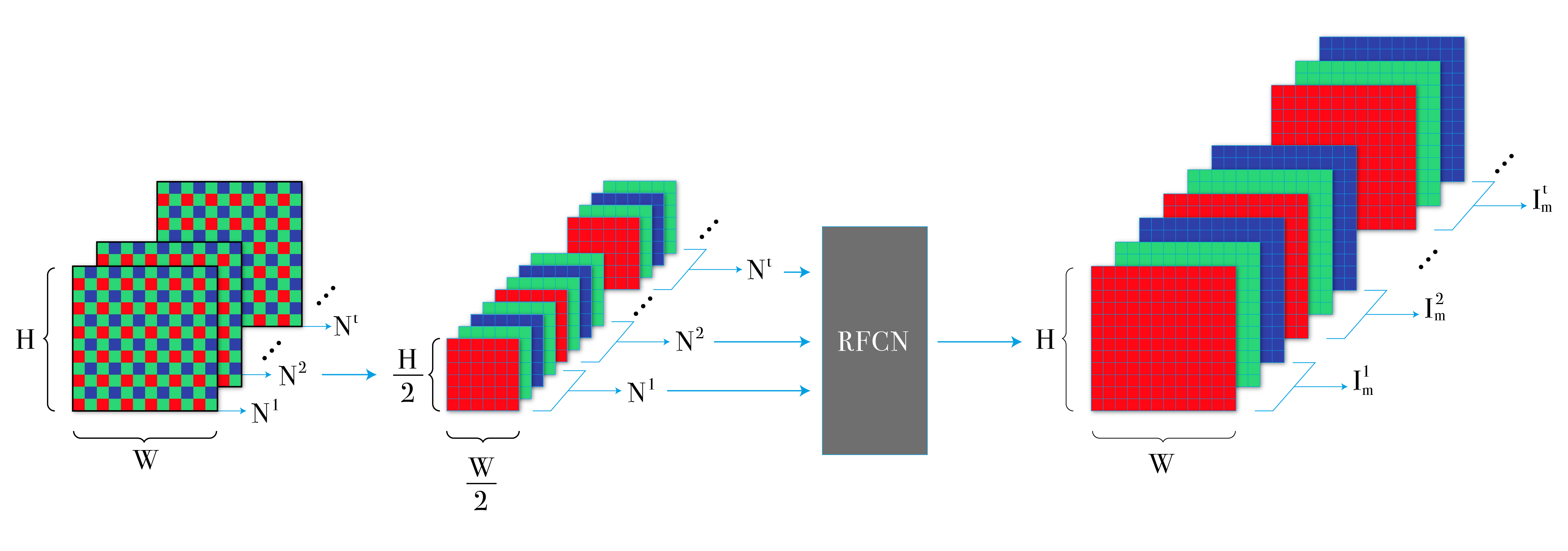}
\caption{A conceptual illustration of the system framework. }
\label{fig1}
\end{figure*}

\section{Related Work}

Image denoising is a long-established low-level computer vision task. Many traditional methods have been proposed to solve the denoising problem, such as nonlocal filtering, Block-Matching and 3D Filtering (BM3D), Weighted Nuclear Norm Minimization (WNNM), and so on~\cite{buades2005non}\cite{dabov2007video}\cite{gu2014weighted}. Meanwhile, computer vision has made rapid progress with the development of deep neural networks~\cite{remez2017deep}. One of the first attempts to use deep neural networks for image denosing was DnCNN~\cite{zhang2017beyond}. DnCNN can realize end-to-end denoising by using neural networks, as well as by adopting residual learning strategies which can achieve blind image denoising, and largely surpasses traditional methods. Residual Encoder-Decoder networks (RED) propose a residual network with skip connections~\cite{chen2017low}. RED uses a  convolution and deconvolution symmetric network structure. Convolution operations extract features, while deconvolution is an upsampling of the extracted features; thus, the technique completes the whole procedure from image to feature, and then from feature to image. Similar to RED, U-net has also been used for image denoising and has achieved good results~\cite{ronneberger2015u}. U-net also has skipping connections, and its large receptive field is capable of effectively reducing the number of layers. We also use U-Net-like Fully Convolutional Network (FCN) architecture in this work.

 Some multi-frame denoising methods have also been proposed and can usually achieve better results than single-image denoising~\cite{hasinoff2016burst}. The fusion of multi-frame images can effectively improve SNR, hence enhancing the image quality. Among the traditional multi-frame denoising methods, V-BM4D finds the motion trajectory of the target block between the frames, regards the series of in-motion trajectories as the target volume, and then finds similar volumes to form a four-dimensional set. Finally, it performs filtering in the four-dimensional set, which can achieve effective video denoising~\cite{maggioni2011video}. There have been several attempts to deal with multi-frame image denoising through deep neural networks. Godard, et. al., uses a Recurrent Neural Network (RNN) ~\cite{godard2018deep} . The use of RNN can efficiently aggregate the information of the frames before and after, as well as increasing the effective depth of the network to enlarge the receptive field~\cite{mikolov2010recurrent}. It is worth noting that ~\cite{godard2018deep} does not use skip connections. We also use RNN, but with skip connections to combine multi-frame image information for denoising and enhancement.

In the case of extremely dark environments, Chen, et. al., proposed a new pipeline to replace the traditional one, which includes several procedures including balance, demosaicing, denoising, sharpening, color space conversion, gamma correction, and more. These processes are specifically tuned in the camera module to suit the hardware. However, because of these non-linear processes on the raw data, some information is lost. Starting from raw data can help to improve the image quality. Previous research has proven that using raw data instead of sRGB can effectively enhance the quality of denoising. Raw images are generally 10bit, 12bit or 14bit, and often contain more bits of information than 8-bit sRGB images~\cite{schwartz2019deepisp}. Especially in the case of extremely dark environments, raw image can be used to obtain more low-brightness information. Therefore, we will use an end to end system starting from raw data and directly generating an output of a denoised and enhanced sRGB image. We handed over all the processes that were originally handled by the ISP to the neural network. On the one hand, the neural network can fully use the information of these raw images, and on the other hand, it will simplify the pipeline.

\begin{figure*}[t]
\centering
\includegraphics[width=14cm]{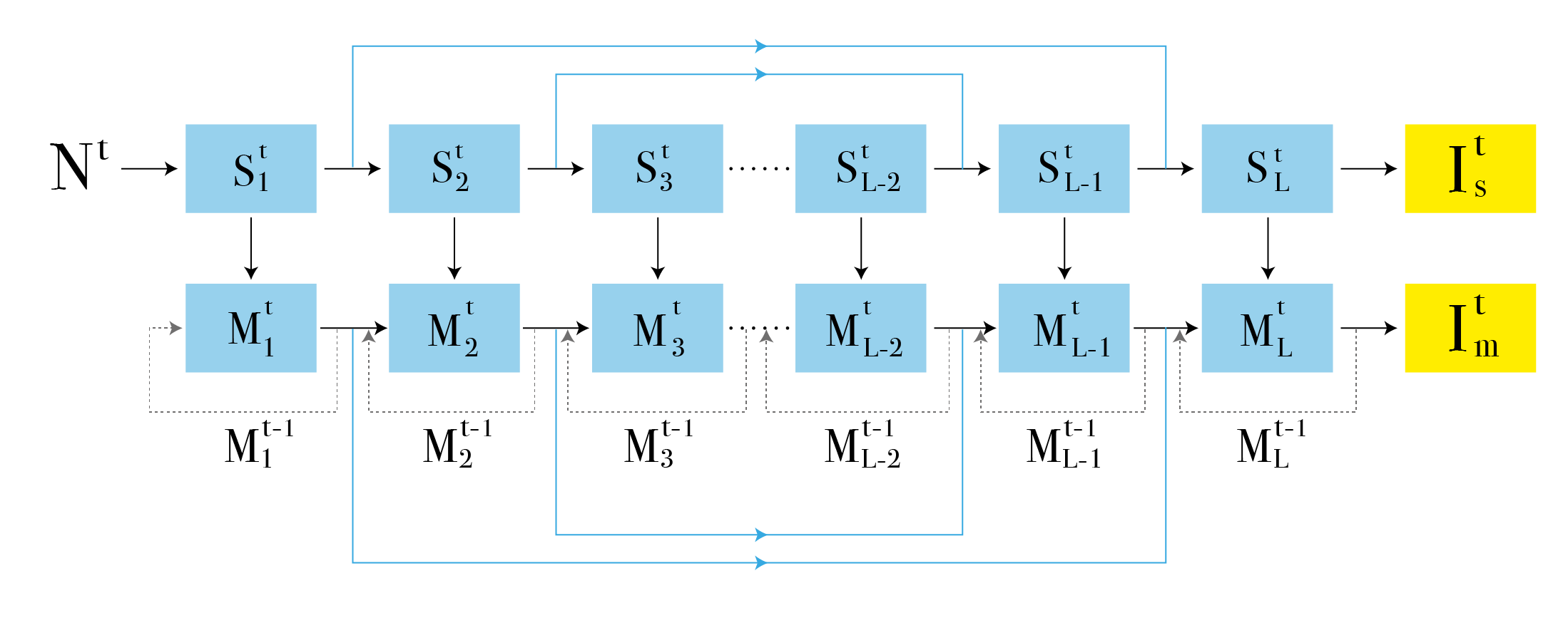}
\caption{This image illustrates the network architecture, in which Frame t is taken as an example. The above part is a single-frame network from $S_1$ to $S_L$, in which L is the number of the layers. It is a U-Net structure with skip connections. The multi-frame recurrent network from $M_1$ to $M_L$, shown below, is also a U-Net structure with skip connections. There is a recurrent connection in each unit of the multi-frame network. The multi-frame network takes each scale feature of the single-frame network as an input at each recurrent connection, and the output obtained by the unit in the previous frame is also concatenated to this unit. In contrast to common RNNs, convolution is used in place of Gated Recurrent Units (GRU) and Long Short Term Memory (LSTM). Finally, $I_{s}^{t}$ represents output from single-frame network for Frame t with $I_{m}^{t}$ representing output from multi-frame network for Frame t. They will be used for calculating the loss. $I_{m}^{t}$ is also the denoised output image of Frame t.}
\label{fig2}
\end{figure*}

\section{Methods and Materials}

In the following sections, we describe how to implement the proposed network for multi-frame denoising.

\subsection{Framework}

Fig. 1 represents the framework of our system.  After obtaining the raw burst data, we organize multi-frame raw images into appropriate structures as input for the neural network. For the Bayer Pattern raw image, a common practice is to pack 4 color channels with the resolution of each channel reduced by half. Then, we subtract the black level, which is the value produced by the dark current released by the photodiode without light. Most of the raw image values are distributed in areas close to the black level in extremely dark environments; different cameras produce different black level values. Subtracting black level from the raw image can help to apply the trained model directly to raw images from different cameras. This linear processing does not affect valid information. Following this, we scale the data according to the desired amplification ratio, which is a factor that can be tuned ~\cite{chen2018learning}. The amplification of brightness is difficult to acquire in the convolutional neural network denoising task, especially in different raw images. Instead of using neural networks to acquire this coefficient,  it is more appropriate to set a separate amplification ratio outside of the network. By tuning the amplification ratio, we can satisfy the different needs of a variety of scenarios. When training the model, we need to multiply the amplification ratio to get an appropriate input brightness to match the brightness of the ground-truth. The packed and amplified data is then fed into the RFCN. The RFCN network has an architecture which is composed of a U-net combined with RNN. Finally, the network generates a multi-frame denoised image sequence.

\subsection{Network Architecture}

Fig. 2 shows the network architecture. We propose to use the RNN method to fully process multi-frame images; the multi-scale  features can be fused in a recurrent manner to obtain context information and perform sequential processing. As a network for processing sequential data, RNN is relatively flexible and  easy-to-expand. For our entire network, all parameters are shared, with each frame using the same parameters. Parameter sharing enables a reduction in the number of parameters, which results in a shortened training time and a reduced possibility of overfitting. For CNN, parameter sharing is cross-regional, while for RNN, parameter sharing is cross-sequence; thus, parameters are shared by deep computational graphs. The convolution kernel for any position in any frame is the same. Therefore, the entire network can be extended to sequences of any length. 

Similar to the technique used in ~\cite{godard2018deep} , we use the overall network architecture of a dual network, which is divided into a single-frame network and a multi-frame RNN network. The single-frame network is U-Net, which is a fully convolutional neural network that is suitable for any size of input. U-net's different scale features are fed separately to the corresponding scale recurrent connections in the multi-frame network.  These features are preliminary processed feature information and are more efficient for multi-frame networks. The multi-frame network takes each scale feature of the single-frame network as input at each recurrent connection. The single-frame network first processes each frame of image separately, and then inputs the multi-scale features into the multi-frame recurrent network. Since the basic structure we use is U-net, for single-frame networks and multi-frame networks the process from the front to the back of the network is downsampling and upsampling. Moreover, the single-frame network and the multi-frame network are consistently sampled up and down, thereby ensuring that the scale of the features is consistent. Compared to the structure used in ~\cite{godard2018deep}, we can use the U-net to better extract information. 

For the entire network, there are F frames of raw images as input and F frames sRGB of images as output. The latter output frame indirectly utilizes all of the previous information. More output information can be aggregated in the later frame. It is equivalent to a very deep network. In general, the later the frame, the more denoising has been performed, and therefore the higher the image quality obtained.

\begin{figure*}[h]
\centering
\includegraphics[width=16cm]{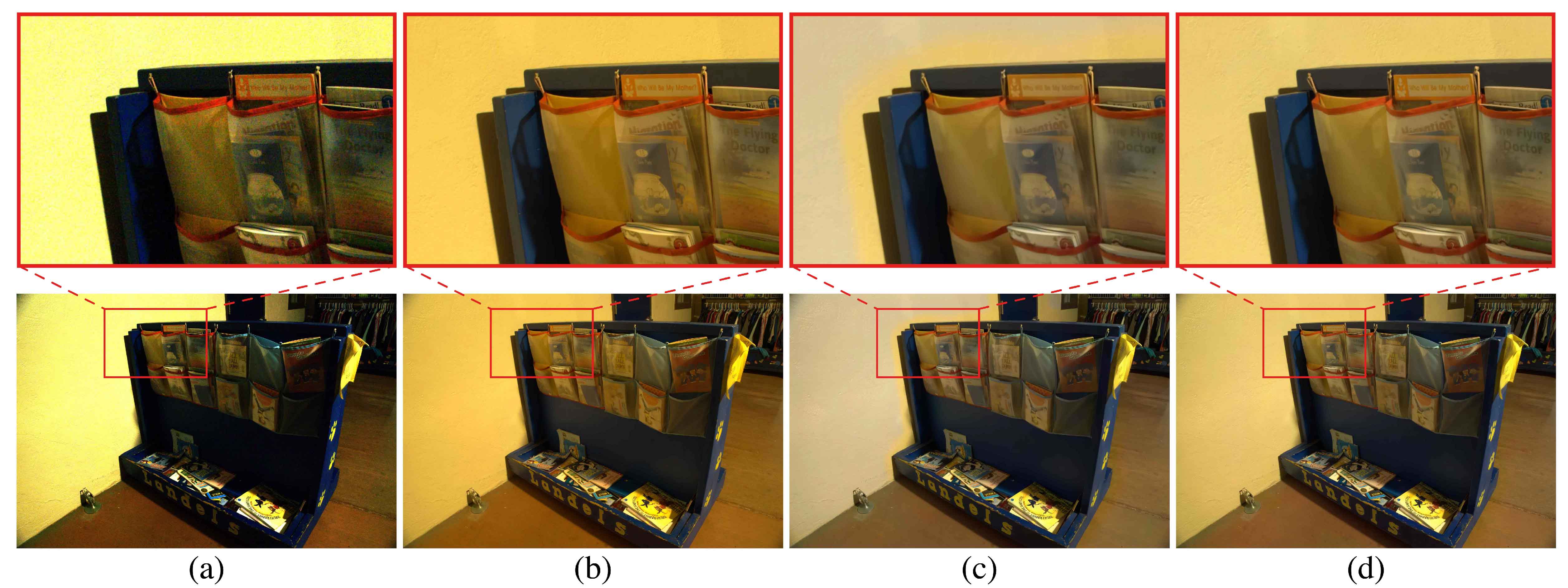}
\caption{Four images: (a) was produced using the traditional image processing pipeline; (b) is the ground-truth, taken using long exposure; (c) was produced using a single-frame enhanced network, as per ~\cite{chen2018learning}; (d) was produced using our multi-frame network. It can be clearly seen from the enlarged portion of (c) as compared with (a) that (c) is less noisy; however, the wall color is uneven and does not correspond to the ground-truth and, furthermore, the details of the magazine covers are lacking. In contrast, in (d) the enlarged portion shows that the wall color is much closer than (c) to ground-truth, and more details have been recovered. The PSNR and SSIM of (c) are 21.82 and 0.889 respectively, while the PSNR and SSIM of (d) are 24.93 and 0.903 respectively.}
\label{fig2}
\end{figure*}

\subsection{Data}
There are very few datasets for extremely dark environments. The most related one is the See-in-the-Dark (SID) dataset~\cite{chen2018learning}. This dataset contains different sets of raw short-exposure burst images. Each set has a corresponding long-exposure reference image which captures the same scene, which will be used as ground-truth. They were all obtained by real life phone camera shots under very dark environments. Outdoor photos were taken at night under moonlight or streetlights, with illumination ranging from 0.2 lux to 5 lux; and indoor photos' illumination ranges from 0.03 lux to 0.3 lux. Therefore, the images were taken under extremely dark environments, but nothing out of the ordinary for normal cell phone camera users. Each group contains up to 3 types of short exposure bursts of different exposure times: 0.033s, 0.04s, and 0.1s, respectively. The corresponding ground-truth is a long exposure image. Its exposure time is 10s or 30s; although there is still some noise, the quality can be considered high enough. All photos were remotely controlled by a tripod and did not require alignment. We chose the Sony camera Bayer pattern raw images with a resolution of $4240*2832$ as our main training datasets. In addition, we collected data from extremely dark situations taken by other mobile phones as a generalized test of trained models.

\subsection{Training Details}
We used the TensorFlow framework. The single-frame denoising network  regresses  denoised image $I_{s}^{t}=f_{s}(N^{t}, \theta_{s})$ from noisy raw input $N^{t}$, given the model parameters $\theta_{s}$, while the multi-frame denoising network regresses each noisy frame, $I_{m}^{t}=f_{m}^{t}({N^{t}}, \theta_{m})$, given the model parameters $\theta_{m}$. We train the network by minimizing the L1 distance between the predicted outputs and the ground-truth target images as follows~\cite{godard2018deep}:
\begin{equation*}
E=\sum_{t}^{F} \left| I^{t}-f_{s}(N^{t}, \theta_{s} )\right| +  \left| I^{t}-f_{m}^{t}([N^{t}], \theta_{m} )\right| 
\end{equation*}

The patch size is 512*512. We used a relatively large patch size because, when using U-Net, the image quality of the patch edge is not as good as the middle after downsampling and upsampling. A large patch size ensures that the training would be satisfactory. We also performed data augmentation on the dataset. We set the Adam with learning rate to $0.5*10^{-4}$, then decay it by one half every 1000 epochs. 137 sequences (10 burst images, $4240* 2832$) were used for training, while 41 were reserved for testing.

\begin{figure*}
\centering
\includegraphics[width=14cm]{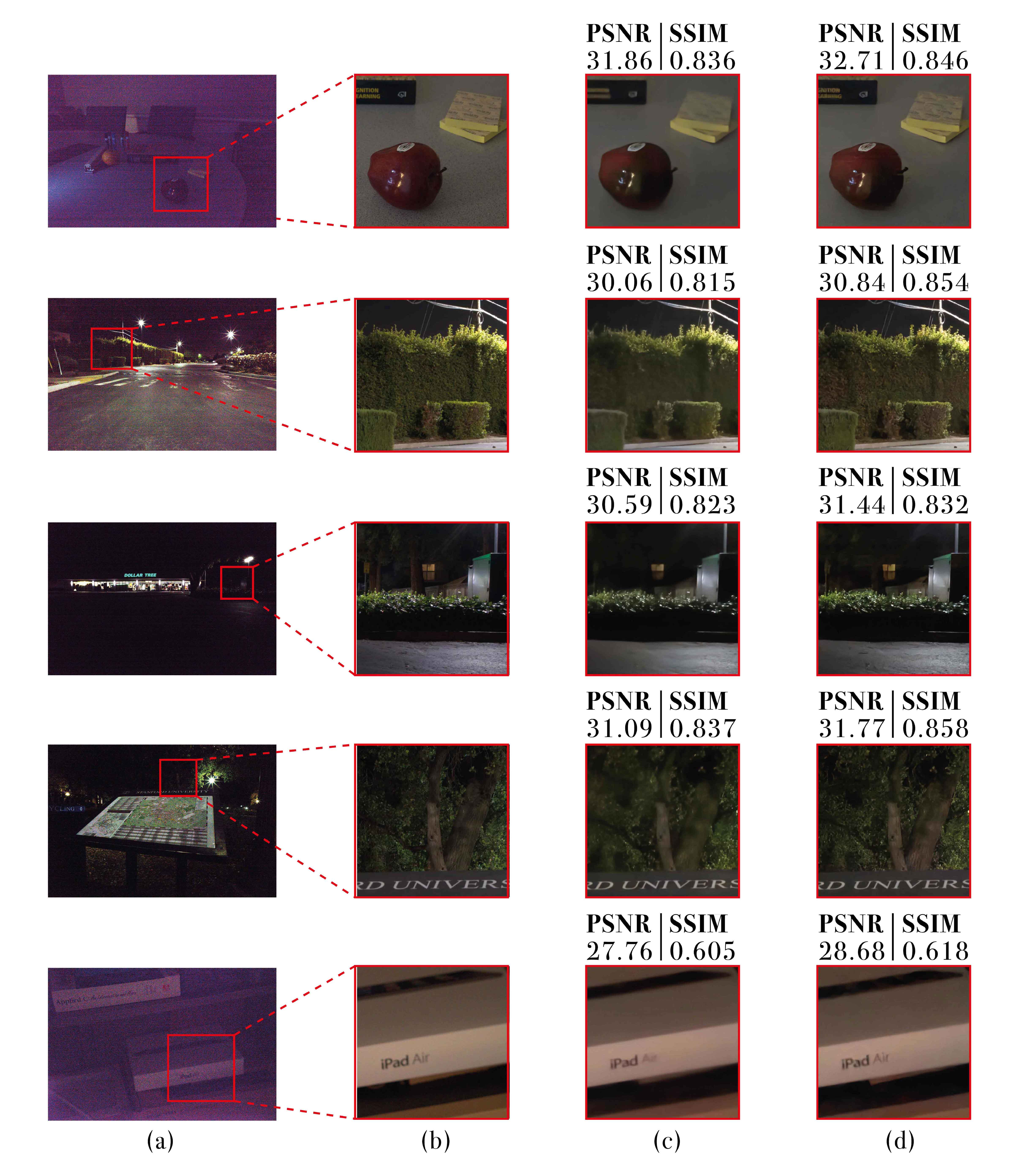}
\caption{In Column (a), the images were obtained via a traditional image processing pipeline. In Column (b), the images are the ground-truth, as obtained by long-exposure. In Column (c), the images represent the baseline for fair comparison, which is calculated by inputting each burst into ~\cite{chen2018learning} network for denoising, after which the output is averaged. In Column (d), the images represent our results. As can be seen, in Column (d), the colors and details are more accurate and correspond better to the ground-truth as shown in Column (b). This can best be seen on a screen, where the images can be magnified.}
\label{fig2}
\end{figure*}

\begin{figure}
\centering
\includegraphics[width=8cm]{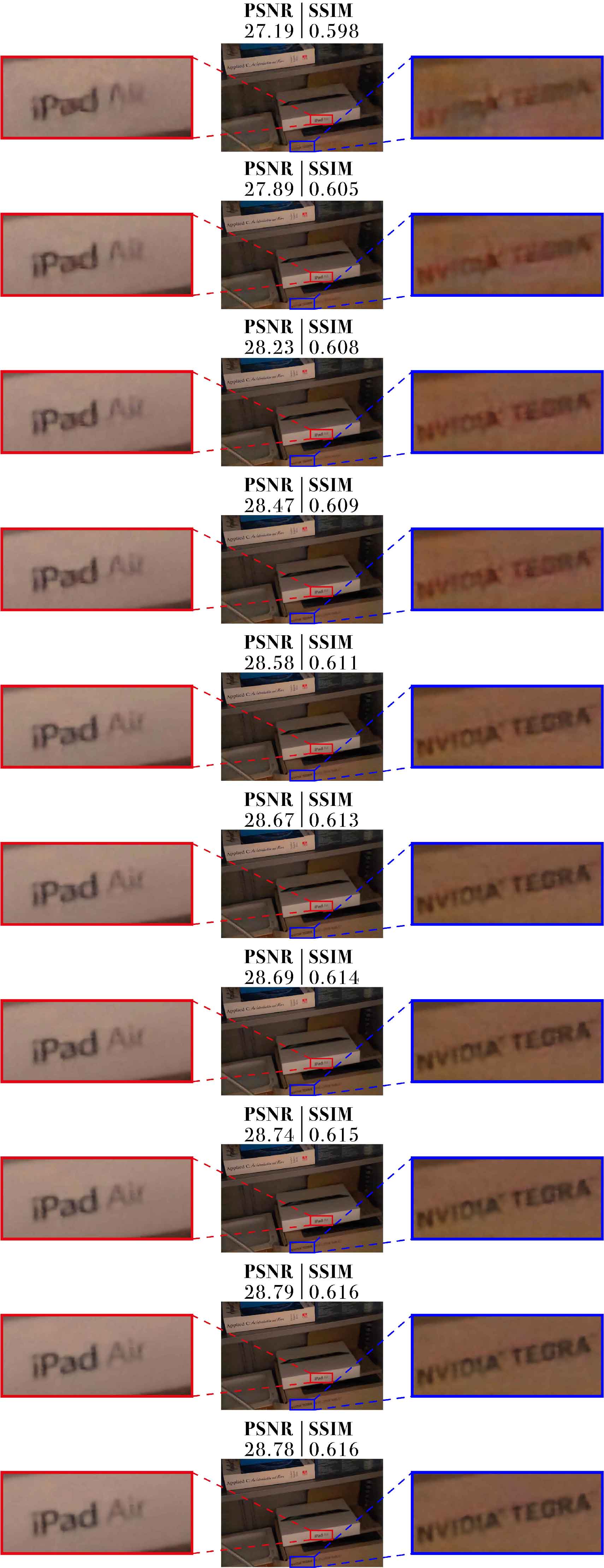}
\caption{This figure shows the results of a 10-frame output model, arranged from the first frame (top) to the last (bottom). PSNR and SSIM generally increase in value over these ten progressive frames; this can be confirmed visually as well, with the text on the frame image generally becoming clearer in each successive frame.}
\label{fig2}
\end{figure}

\begin{figure}
\centering
\includegraphics[width=8cm]{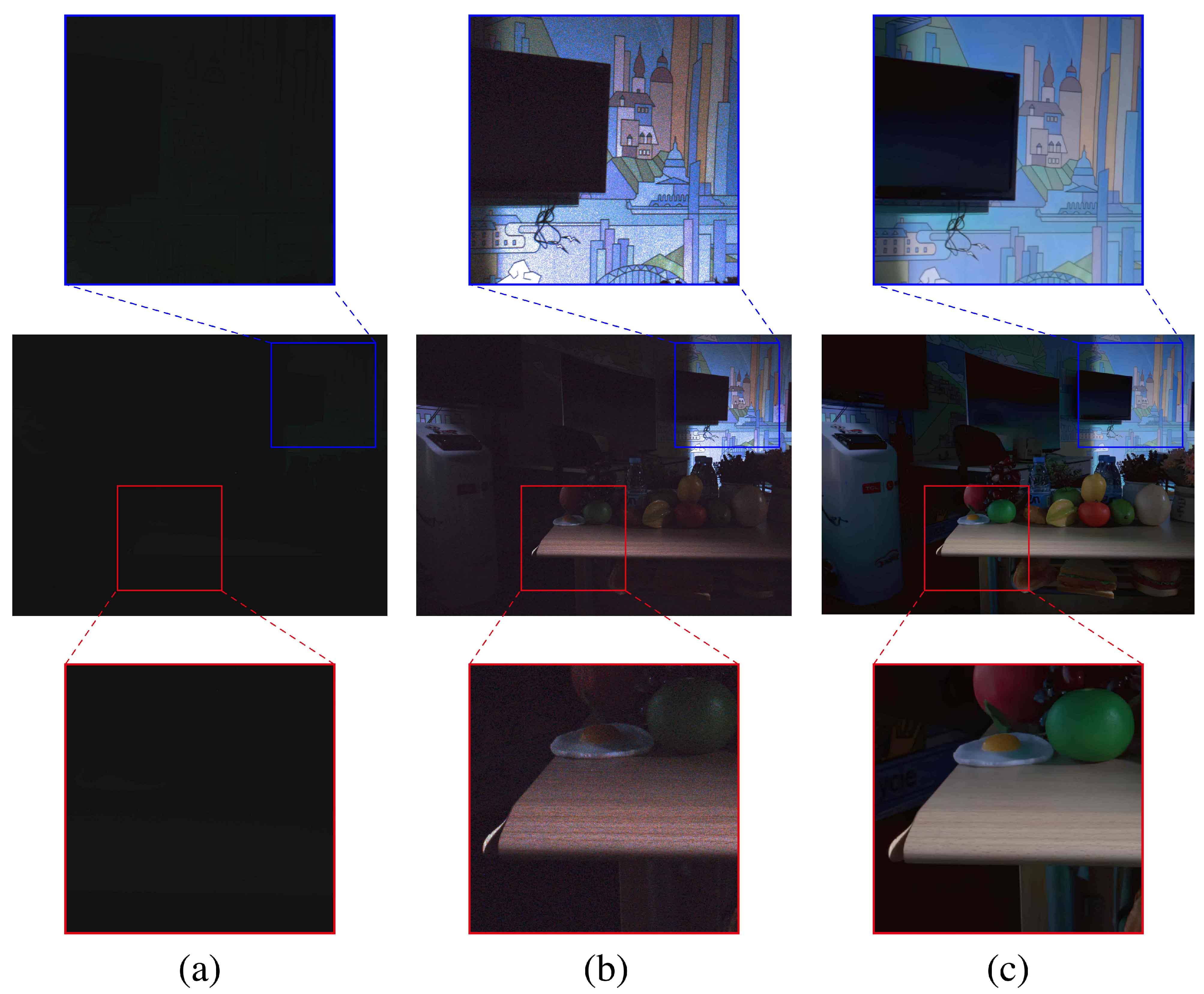}
\caption{This example shows how the trained model applies to different cameras' raw bursts (Blackberry key2, 10 bursts, exposure time: 0.1s). In (a), we converted the original raw burst data directly using traditional pipeline to an RGB image. (b) Due to excessive darkness, we use Photoshop to brighten the image in order to see the content. The detailed image clearly has a lot of noise. In (c), we applied the trained model to the bursts. From the images it can be seen that the trained model can be applied to different cameras' images and can obtain good denoising and enhancement.}
\label{fig5}
\end{figure}

\begin{figure}
\centering
\includegraphics[width=7.5cm]{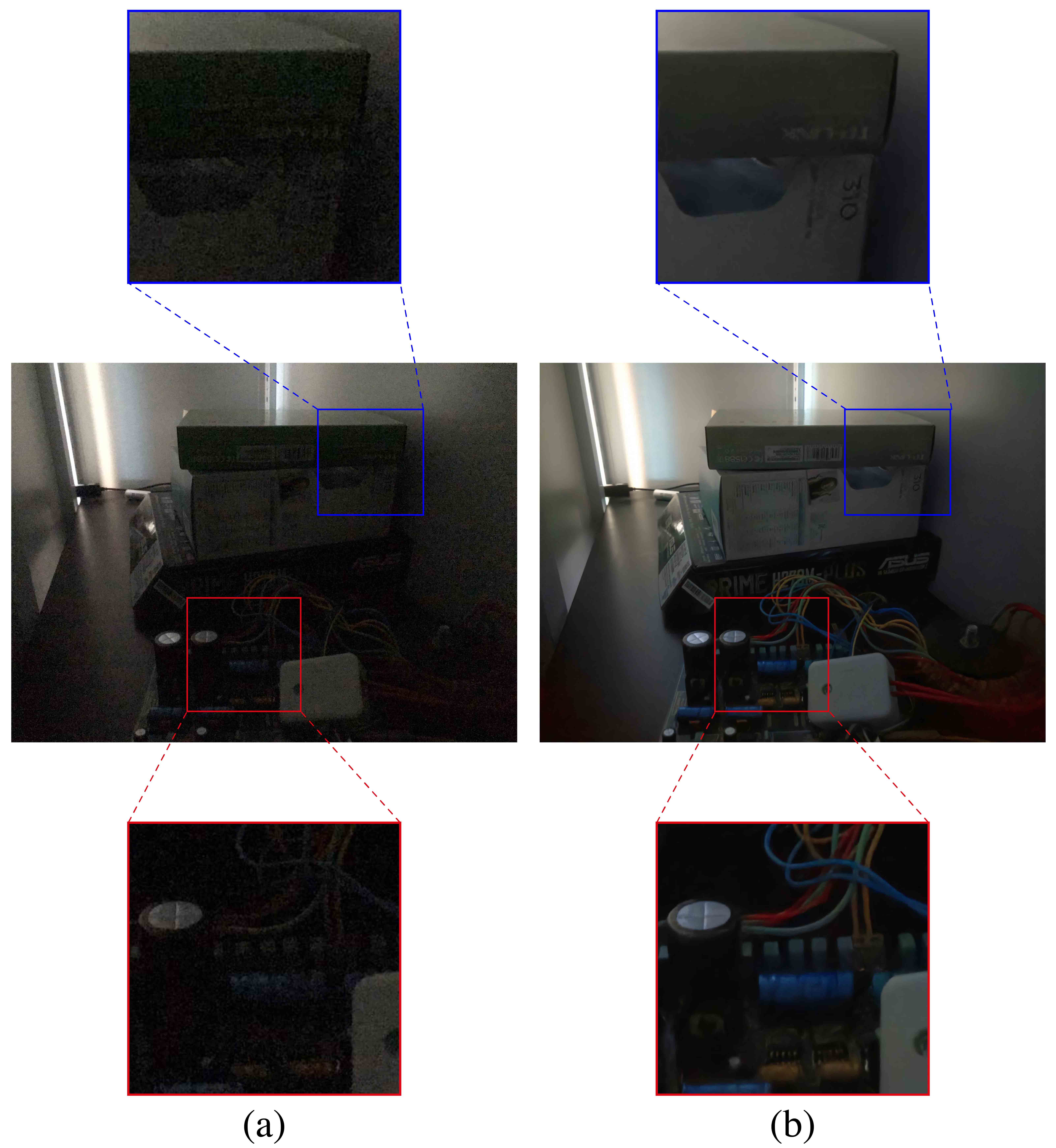}
\caption{This is a second example which shows how the trained model applies to different cameras' raw bursts (iPhone 7 Plus, 10 bursts, exposure time: 0.033s). In (a), we converted one of the original raw burst data directly using traditional pipeline to an RGB image. In (b), we  applied the trained model to the bursts. From the images it can be seen that the trained model can be applied to different cameras' images and can obtain good denoising and enhancement.}
\label{fig5} 
\end{figure}

\section{Results and Discussion}

To begin with, we use one example from the SID dataset to compare three images to the ground truth: first, the image produced by the traditional image processing pipeline; second, the image produced by ~\cite{chen2018learning}, which is a single-frame enhanced network; and third, our result. We can see from Fig. 3 that, compared with the traditional image processing pipeline-generated image, the single-frame enhanced network has already greatly improved the image, with greatly reduced noise when compared with ground-truth (the long exposure image). However, the result still lacks many details; additionally, sometimes the colors in the image do not reflect the real color. This is due to the fact that the SNR of the shot's image is quite low under dark environment. Strong noise conceals a substantial amount of useful information, which cannot be recovered completely through a single image. In contrast, use of our multi-frame network permits more effecitive recovery of the details and corrects the colors to make them more in line with the ground-truth.

For fair comparison, each frame of the 10 bursts of the same scene in the SID dataset was input to the network for denoising and then averaged. This works as a baseline for comparison. We compared our 10-frame denoised results with the baseline and found that our network can obtain more details, less noise, and better enhancement. Some of the resulting images can be seen in Fig. 4. As is clear from these images, in the first example, the average processed image cannot clearly identify the writing on the sticky note, while the images obtained by our network are significantly closer to the ground-truth. Furthermore, the color of the apple in our image is also closer to the ground-truth. Similarly, in the second row, the images obtained by our network clearly show the details of the leaves, while the leaves in the average processed image are blurred. Continuing to the third row, the image obtained by our network reveals the outline of the house and the texture of the ground, which cannot be seen in the average processed image. In the fourth row, the image obtained by our network shows the texture of the tree’s trunk and branches, while in the average processed image these things cannot be seen clearly. Finally, in the fifth row, the text in the image obtained by our network is clearer than the text in the average processed image. We have calculated the corresponding Peak Signal-to-Noise Ratio (PSNR) and Structural Similarity (SSIM). On average, the PSNR and SSIM of our results are 30.75 and 0.822 respectively, while the PSNR and SSIM of the baseline are 30.06 and 0.808 respectively. The PSNR of our results can exceed 0.69dB, while the SSIM has increased by 0.014 from the baseline.

To explore how our network processes image sequences, we trained a 10-frame model and output the first to last frames of the image on the test set. Figure 5 shows the results of the 10 output frames. PSNR and SSIM generally show an increasing trend in value, while the text in the image of each frame can be visually confirmed as becoming less blurry and clearer in each successive frame. This is because our recurrent architecture processes the image sequence frame-by-frame, with each successive frame aggregating and utilizing all of the previous information from each preceding frame. In general, the more previous information which is aggregated and utilized, the better the later output will be.

We also collected some RAW images taken with other mobile phones. These images, too, were taken in very dark environments. Fig.6 and Fig.7 show how the trained model applies to different cameras' raw bursts. We used our trained model to process these images and obtained the same good results without fine-tuning. In theory, because of the different sensors, these images could be used as datasets to train and retest in order to achieve the best results. However, we directly used the model that was trained with SID before and tested the different camera's datasets, which led to good results and makes the photos similar to the ones in the previous experiments. This shows that our network can be generalized between different models of cameras. The model obtained by training images captured by a camera can be effectively used on images obtained by other cameras, and thus has good portability. It is only necessary to repack the Bayer pattern of the model according to the Bayer pattern of the corresponding camera, putting it in the same order. The black level of the corresponding camera is subtracted in the normalization, thus allowing completion of the migration.

\section{Conclusions}

The proposed framework, based on an RFCN (i.e., a U-net combined with RNN), is designed to process raw burst photos taken under extremely low-light conditions and to obtain denoised images with improved brightness. All of this is end-to-end processing done through our network. We have illustrated that the use of raw bursts images obtains better results than state-of-the-art methods under dark environments. Additionally, our model maps raw burst images directly to sRGB outputs, either producing a best image or generating a multi-frame denoised image sequence. As a consequence, our framework has a relatively high level of flexibility, and opens up the possibility of expanding our framework to cover video as well as image denoising. Finally, we have proven that our framework is highly portable with a great deal of cross-platform potential; therefore, the model trained by one mobile phone can be directly applied to another camera's raw bursts without the necessity fine-tuning, and a similar level of enhancement can be expected. 

In the future, by optimizing the network architecture and training procedure, we expect to continue to yield further improvements in image quality. 

{\small
\bibliographystyle{ieee}
\bibliography{egbib}
}

\end{document}